\documentclass[10pt,twocolumn,letterpaper]{article}

\usepackage{cvpr}              
\usepackage{bm}
\usepackage{float}
\usepackage{algorithm}
\usepackage{algorithmic}

\newcommand{\myparagraph}[1]{\vspace{0mm}\noindent\emph{\textbf{#1}}}

\definecolor{cvprblue}{rgb}{0.21,0.49,0.74}

\usepackage[pagebackref,breaklinks,colorlinks,allcolors=cvprblue]{hyperref}

\def\paperID{354} 
\def\confName{CVPR}
\def\confYear{2025}

\def\sysname{DAGSM}
\def\modelname{GSM}

\title{DAGSM: Disentangled Avatar Generation with GS-enhanced Mesh}

\author{
Jingyu Zhuang\textsuperscript{1,2}
\quad 
Di Kang\textsuperscript{2}
\quad
Linchao Bao\textsuperscript{2}
\quad
Liang Lin\textsuperscript{1,3,4}\footnotemark[1]
\quad
Guanbin Li\textsuperscript{1,3,4}
\\
\textsuperscript{1}Sun Yat-sen University \qquad \textsuperscript{2}Tencent \qquad \textsuperscript{3} Peng Cheng Laboratory \\
\textsuperscript{4}Guangdong Key Laboratory of Big Data Analysis and Processing  \\
{\tt\small zhuangjy6@mail2.sysu.edu.cn, di.kang@outlook.com, linchaobao@gmail.com,} \\
{\tt\small linliang@ieee.org, liguanbin@mail.sysu.edu.cn}
}

\begin{document}

\setlength{\abovedisplayskip}{5pt}
\setlength{\belowdisplayskip}{5pt}
\setlength{\textfloatsep}{6pt}
\setlength{\dbltextfloatsep}{6pt}  
\setlength{\floatsep}{4pt}
\setlength{\dblfloatsep}{4pt}  
\setlength{\intextsep}{4pt}
\setlength{\abovecaptionskip}{4pt}
\setlength{\belowcaptionskip}{4pt}

\twocolumn[{%
\renewcommand\twocolumn[1][]{#1}%
\maketitle
\vspace{-20pt}
\begin{center}
    \centering
    \captionsetup{type=figure}
    \includegraphics[width=0.97\textwidth]{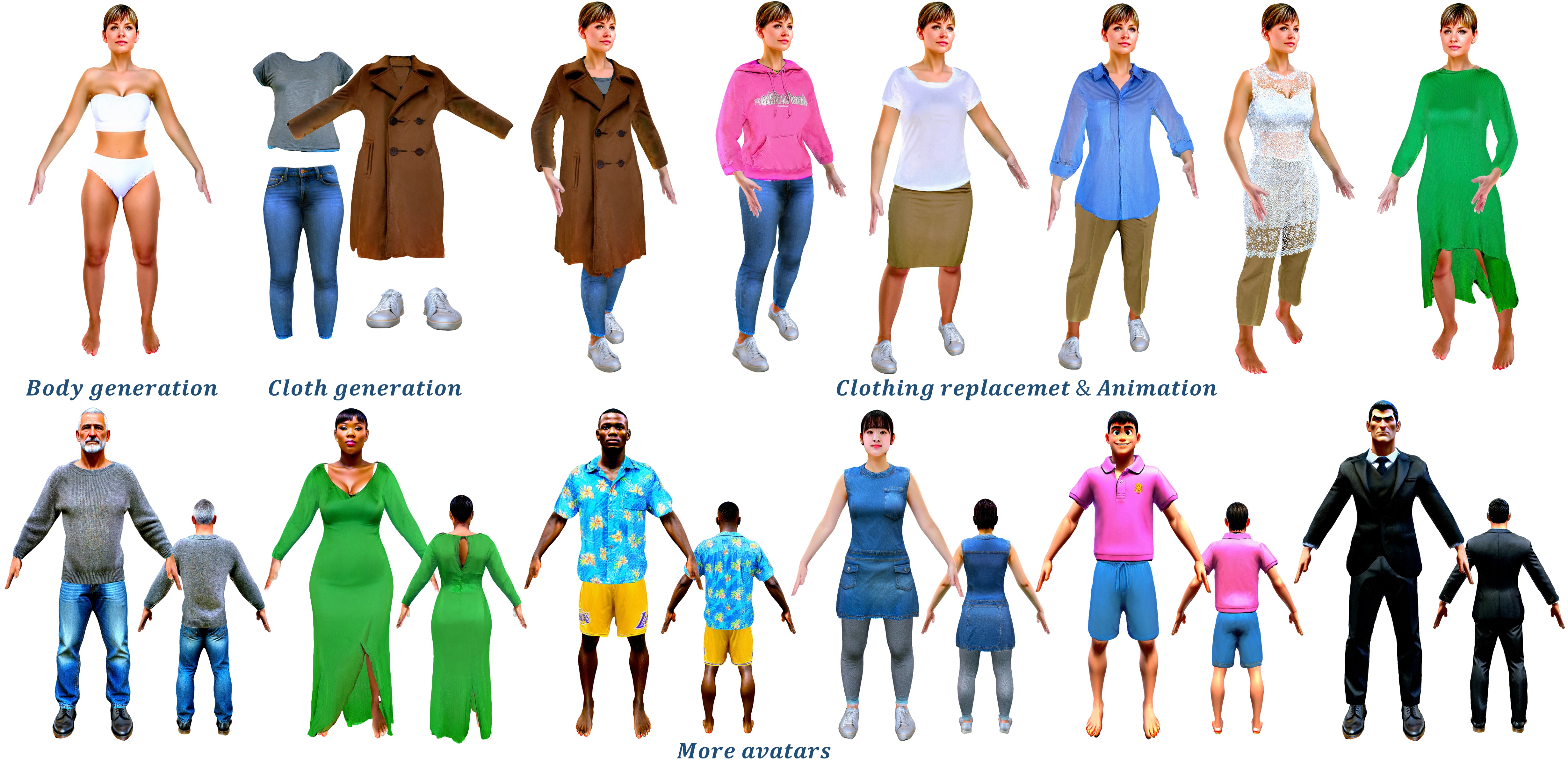}
    \captionof{figure}
{
Given text prompts, our method \sysname{} allows the users to generate disentangled avatars in diverse styles  (e.g., real, cartoon) with various garments.
Our method separately generates the human body and garments for disentanglement, so our method naturally supports clothing replacement.
We represent every single part (e.g., body, upper/lower clothes) using a hybrid GS-enhanced mesh, where the 2D Gaussians are attached on a proxy mesh to better handle complicated cloth texture (e.g., cotton, woolen, and transparent fabric in row 1, left 2) and produce realistic animations.
}
  \label{fig:teaser}
\end{center}%
}]

\maketitle

\renewcommand{\thefootnote}{$\ast$}
\footnotetext[2]{Corresponding author is Liang Lin. Welcome to \href{https://zjy526223908.github.io/DAGSM/}{\textcolor{red}{\emph{Project page}}}}

\begin{abstract}

Text-driven avatar generation has gained significant attention owing to its convenience. 
However, existing methods typically model the human body with all garments as a \emph{single} 3D model, limiting its usability, such as clothing replacement, and reducing user control over the generation process.
To overcome the limitations above, we propose \sysname{}, a novel pipeline that generates disentangled human bodies and garments from the given text prompts.
Specifically, we model each part (e.g., body, upper/lower clothes) of the clothed human as one GS-enhanced mesh (GSM), which is a traditional mesh attached with 2D Gaussians to better handle complicated textures (e.g., woolen, translucent clothes) and produce realistic cloth animations.
During the generation, we first create the unclothed body, followed by a sequence of individual cloth generation based on the body, where we introduce a semantic-based algorithm to achieve better human-cloth and garment-garment separation.
To improve texture quality, we propose a view-consistent texture refinement module, 
including a cross-view attention mechanism for texture style consistency and an incident-angle-weighted denoising (IAW-DE) strategy to update the appearance.
Extensive experiments have demonstrated that \sysname{} generates high-quality disentangled avatars, supports clothing replacement and realistic animation, and outperforms the baselines in visual quality.

\end{abstract}

\vspace{-10pt}
\section{Introduction}

High-quality digital humans are essential for many applications, such as VR, immersive telepresence, virtual try-ons, and video games. 
For better interaction and immersion, it would be necessary to provide users with animatable high-quality avatars that support customizable and flexible clothes combinations according to user wishes.

Although avatars created following the traditional pipeline fulfill all the aforementioned requirements, they involve too much manual work and are thus very time-consuming and expensive.
The industry demands a faster and easier avatar creation pipeline to make it available for both professional users and novices.
Recently, DreamFusion~\cite{poole2022dreamfusion} proposed SDS loss to enable text-to-3D generation based on text-to-image diffusion models, significantly reducing the complexity of generating 3D models.
Based on SDS, dedicated human generation methods~\cite{hong2022avatarclip,kolotouros2024dreamhuman,cao2024dreamavatar} usually introduce parameter human models (e.g., SMPL~\cite{pavlakos2019expressive}) to improve the geometry quality and, most importantly, to animate the generated human model.

However, these methods have two major drawbacks that significantly limit the application of the generated 3D models.
First, these methods generate a \emph{single} model containing both the body and all garments in a single-stage optimization, which prevents clothing replacement, results in unrealistic animation due to clothes adhering to the body, and reduces user control over complex garment combinations.
This significantly limits the application of the generated avatars in virtual try-on and video games.
Second, the texture generated directly by SDS is usually of low quality (e.g., overly smoothed, over-saturated colors), reducing visual appeal and immersion in the user experience.

To alleviate the above issues, we present \sysname{}, which sequentially generates the human body and various garments as individual models represented by GS-enhanced meshes, enabling flexible garment combinations and easier editing.
Due to the involvement of meshes, we can utilize physical simulation to animate the generated avatars and clothes, significantly improving the realism.
Specifically, our method achieves such capabilities through three crucial designs.
(1) A sequential generation pipeline that first generates the unclothed human body, followed by garment generation (conditioned on the human body) and texture refinement.
(2) A GS-enhanced mesh (\modelname{})
representation that binds 2DGS~\cite{huang20242d} onto a mesh to guide the deformation of the Gaussians in animation.
Naturally, we can utilize cloth simulation techniques to obtain more convincing animations.
(3) Two crucial techniques enabling flexible garment combinations and high-quality texture: SAM-based filtering for better cloth separation (i.e., clearer body-garment and garment-garment boundaries) and view-consistent refinement with a cross-view attention mechanism for enhancing texture quality.

We conduct extensive experiments, demonstrating that \sysname{} generates higher-quality avatars with disentangled garments and supports more features, including clothing replacement, manual texture editing, and realistic animation.
As illustrated in Fig.~\ref{fig:teaser} and Fig.~\ref{fig:visual_compare}, \sysname{} supports clothing replacement and realistic animation with natural clothing motion.
Moreover, our method allows precise appearance control by providing a reference image (Fig.~\ref{fig_app_ref}).

Our contributions are summarized as follows:
\begin{itemize}

\item We present \sysname{}, a text-driven framework that includes sequential body/cloth generation and refinement stages to produce high-quality animatable avatars with decoupled bodies and garments.
\item We adopt a hybrid GS-mesh representation (\modelname{}), enabling physical simulation for more realistic animation.
\item We propose two crucial techniques for convincing results: SAM-based filtering for better cloth separation and view-consistent refinement for enhancing texture quality.
\item 
Experiments demonstrate that, compared to the existing method, our \sysname{} generates higher-quality, disentangled avatars with more realistic animations.

\end{itemize}

\section{Related Works}

\subsection{Digital human representations}

Early works~\cite{alldieck2018video,bhatnagar2019multi,ma2020learning} usually combine parametric human meshes (e.g., SMPL~\cite{pavlakos2019expressive})  with vertex offsets to model clothed humans.
PIFu~\cite{saito2019pifu} first introduces the implicit function for modeling, which is widely adopted by many subsequent methods~\cite{saito2020pifuhd,he2020geo,huang2020arch,zheng2021pamir,xiu2022icon,cao2023sesdf,xiu2023econ}. 
With the rapid development of neural radiance fields, many methods~\cite{weng2022humannerf,liu2023hosnerf,peng2021neural,yan2024dialoguenerf} have adopted NeRF~\cite{mildenhall2021nerf} to represent the human body.
Recently, the explicit radiance field 3DGS~\cite{kerbl20233d} quickly draws tremendous attention due to its high rendering quality and efficiency, and some methods~\cite{liu2023animatable,li2023human101,xu2024gaussian,he2025vton360highfidelityvirtual} apply 3DGS in human reconstruction to enhance training efficiency.
Moreover, some methods combine different representations to model the human body.
SelfRecon~\cite{alldieck2018video} combines explicit SMPL+D~\cite{mildenhall2021nerf} and implicit IDR~\cite{yariv2020multiview} to obtain coherent geometry.
DoubleField~\cite{shao2022doublefield} combines Neural Surface Field~\cite{mescheder2019occupancy} and NeRF at the feature level in an implicit manner.
Some methods~\cite{shao2024splattingavatar,zhao2024psavatar,wen2024gomavatar,qian2024gaussianavatars}
use a hybrid representation that binds 3DGS to a single mesh achieving deformation of the avatar. 
However, these methods only reconstruct a single-model clothed human from videos and do not support clothing replacement.

\subsection{Text-guided 3D content generation}

Text-to-3D generation methods, which have been largely advanced by the rapid development of the large Text-to-Image (T2I) generative models, can be categorized into two groups, i.e., feedforward generation and optimization-based generation.
In feedforward generation methods, Brock et al.~\cite{brock2016generative}, SDM-Net~\cite{gao2019sdm}, and SetVAE~\cite{kim2021setvae} train variational autoencoders to generate voxels, meshes, and point clouds, respectively.
Some methods marry GANs~\cite{goodfellow2014generative} with various 3D representations, such as point cloud~\cite{achlioptas2018learning,achlioptas2018learning,shu20193d}, voxel grid~\cite{wu2016learning}, mesh~\cite{cheng2019meshgan}, 
and implicit radiance field~\cite{chan2022efficient, an2023panohead}.
Recently, some methods~\cite{liu2023meshdiffusion,nichol2022point,nichol2022point,jun2023shap,shue20233d,li2023diffusion} propose to train diffusion models for 3D generation.
In optimization-based generation methods, early works use the CLIP~\cite{radford2021learning} to optimize mesh~\cite{michel2022text2mesh,mohammad2022clip} or neural fields~\cite{jain2022zero}.
DreamFusion~\cite{poole2022dreamfusion} proposes score distillation sampling (SDS) loss to distill the knowledge in Text-to-Image diffusion models for text-to-3D generation and editing~\cite{zhuang2023dreameditor,zhuang2024tip}.
Most of the subsequent works~\cite{chen2023fantasia3d,metzer2023latent,lin2023magic3d,wang2023score,wang2024prolificdreamer,chen2024text,tang2023dreamgaussian,yi2023gaussiandreamer} adopt SDS-based optimization pipeline due to its ability to align the generated appearance with the input text.
However, without human body priors, these methods struggle to generate a detailed 3D human body with complex joint structures.

\subsection{Text-guided 3D human generation}

Generating 3D humans rather than generic objects is one most important sub-task and has drawn great attention~\cite{michel2022text2mesh,youwang2022clip,liao2024tada,jiang2023avatarcraft,kolotouros2024dreamhuman,cao2024dreamavatar,huang2024dreamwaltz,liu2024humangaussian,wang2023disentangled,dong2024tela}.
Text2Mesh~\cite{michel2022text2mesh} and CLIP-Actor~\cite{youwang2022clip} use the CLIP model to optimize mesh deformation and generate vertex colors for textures. 
With CLIP loss,  AvatarCLIP~\cite{hong2022avatarclip} and NeRF-Art~\cite{wang2023nerf} adopt  NeRF representation for photo-realistic rendering.
With the emergence of powerful T2I diffusion models, some methods employ SDS loss to optimize 3D human in different representations, such as mesh (TADA~\cite{liao2024tada}), NeRF (Avatarcraft~\cite{jiang2023avatarcraft}, Dreamhuman~\cite{kolotouros2024dreamhuman}, Dreamavatar~\cite{cao2024dreamavatar}, and Dreamwaltz ~\cite{huang2024dreamwaltz}) and 3DGS (HumanGaussian ~\cite{liu2024humangaussian}).
However, the above methods represent the clothed human as a holistic model, making clothing replacement infeasible. 
Recently, a few methods have been proposed to decouple the human body and garments during the generation.
For example, SO-SMPL~\cite{wang2023disentangled} cut off some regions from the body mesh and used them for garment optimization.
However, limited by the predefined mesh topology, it struggles to generate loose clothes, especially those with a different topology to the body (e.g., dresses).
TELA~\cite{dong2024tela} uses multiple NeRFs to represent the body and garments, respectively.
However, the implicit representation hinders garment deformation in animation and reduces animation realism.

\section{Preliminaries}

\noindent \textbf{3D Gaussian Splatting} (3DGS)
~\cite{kerbl20233d} utilizes a set of point-like anisotropic Gaussians $g_i$ to represent a 3D scene: $\mathcal{G} = \{g_1, g_2, ... , g_N\}$.
Each $g_i$ contains a series of optimizable attributes, including center position $\mu$, opacity $\alpha$, 3D covariance matrix $\Sigma$, and color $c$.
The differentiable splatting rendering process is:
\begin{equation}
C =\sum_{i\in \mathcal{N}} c_i \sigma_i \prod_{ i-1}^{j=1}  (1-\sigma_j), \quad \sigma_i= \alpha_i e^{-\frac{1}{2} (x)^{T} \Sigma^{-1}(x) }
\label{eq_gs}
\end{equation}
where $j$ indexes the Gaussians in front of $g_i$ according to their distances to the camera center (i.e. depth),
$\mathcal{N}$ is the number of Gaussians contributed to the ray,
and $c_i$, $ \alpha_i $, and $x_i $ represent the color, density, and distance of the sampling point to the center point of the $ i $-th Gaussian.

In this paper, we employ a variant of 3DGS, i.e. 2DGS~\cite{huang20242d}, which compresses the 3D volume into 2D-oriented Gaussian disks for better surface. 
This approach reduces multi-view inconsistency in 3DGS, enhancing reconstruction quality.

\vspace{5pt}
\noindent \textbf{Optimizing Radiance Fields with RFDS Loss.}
DreamFusion~\cite{poole2022dreamfusion} proposes Score distillation sampling (SDS) to distill the priors from a Text-to-Image (T2I) diffusion model for 3D generation.
\cite{yang2024text} extends this method to the rectified-flow-based T2I models (e.g. Stable Diffusion 3 ~\cite{esser2024scaling}) and proposes RFDS loss:
\begin{equation}
\nabla_{\mathcal{G} }\mathcal{L}_{rfds}(\phi,x)=\mathbb{E}_{\epsilon,t}\bigg[w(t)(x-\epsilon-v_{\phi}(x_{t},y,t))\frac{\partial x}{\partial \mathcal{G}  } \bigg]
\end{equation}
where $w(t)$ represents a positive value related to timestep $t \in [0, 1]$,
$x$ is the image rendered from the radiance field $\mathcal{G}$, $y$ represents the embedding of condition.
The rectified flow network parameter $\phi$ is fixed and $\epsilon$ is the random noise.

\begin{figure*}[t]
\centering
\includegraphics[width=0.95\textwidth]{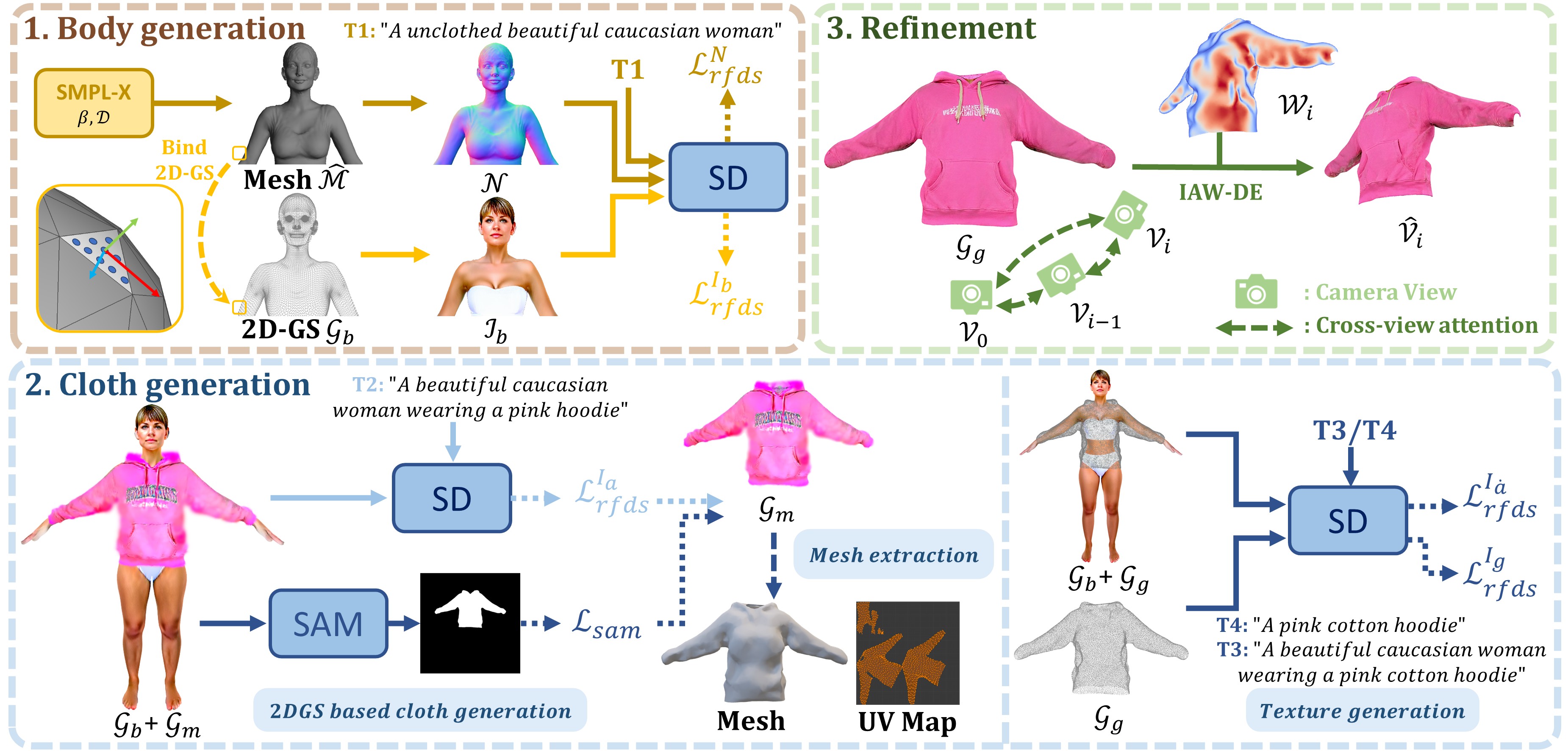}
\caption{
\textbf{Method overview.}
Given text prompts, \sysname{} generates disentangle digital humans whose bodies and clothes are represented as multiple individual \modelname{} (Sec.~\ref{subsec:model}).
The generation process includes three stages: 
1) a body generation stage that generates an unclothed body with the human priors SMPL-X~\cite{pavlakos2019expressive} from the guidance of text-to-image model SD~\cite{esser2024scaling} (Sec.~\ref{subsec:body});
2) a cloth generation stage that first creates the cloth's mesh proxy. 
Then 2DGS $\mathcal{G}_{b}$ is bound to the mesh for generating a garment with texture (Sec.~\ref{subsec:cloth});
and 3) a view-consistent refinement stage, where we propose a cross-view attention mechanism for texture style consistency and an incident-angle-weighted denoising (IAW-DE) strategy to enhance the appearance image $\hat{\mathcal{V}}_i$ (Sec.~\ref{subsec:refinement}).
}
\label{fig_overview}
\end{figure*}

\section{Method}
\label{sec_details}

Given the text prompts describing a human body and the wearing clothes, our goal is to generate a disentangled avatar with high-quality texture, where each garment and the body are decoupled and modeled separately using GS-enhanced meshes (\modelname{}, Sec.~\ref{subsec:model}).

To obtain a disentangled avatar, \sysname{} generates the unclothed human body and garments in different stages, followed by a refinement step (Fig.~\ref{fig_overview}).
\sysname{} contains three major stages: 
(1) We first generate a body in underwear (Sec.~\ref{subsec:body}).
(2) In the following cloth generation stage (Sec.~\ref{subsec:cloth}),
we first generate the mesh proxy of the \modelname{}-based garment, and then optimize the 2DGS of \modelname{} to obtain the garment texture.
Utilizing such a mesh-based representation enables physics-based cloth simulation and makes clothes editing easier.
(3) Finally, we propose a view-consistent refinement stage (Sec.~\ref{subsec:refinement}) to improve the texture quality of the body and garment.

\subsection{\modelname{} representation}
\label{subsec:model}

We propose to represent each model with a GS-enhanced mesh (\modelname{}), 
where the 2DGS are attached to the mesh faces  (Fig.~\ref{fig_overview}, top-left).
The advantages of \modelname{} are two-fold:
(1) significantly enhancing the realism of clothing movement when driving the avatar with the help of a clothes simulator.
(2) enabling easier and more accurate cloth editing via direct texture map modification.

Specifically, the 2DGS are attached to the mesh triangles to follow the mesh transformation during animation/simulation similar to~\cite{qian2024gaussianavatars}.
The 2DGS are defined in a local coordinate system derived from its bind triangle  (Fig.~\ref{fig_overview}, top-left), where we use its centroid position $\bm{T}$ as the origin of the local coordinate.
The direction of one edge, the triangle's normal $\vec{\bm{n}}$, and their cross-product define the three coordinate axes' directions $\bm{R}$.
In the local coordinate system, we define the position of the 2DGS using the barycentric coordinates ($\lambda_1, \lambda_2$) and height offset $z$ along the normal direction (i.e., $\bm{\mu}=\{\lambda_1, \lambda_2,z\}$).
During rendering, the geometry attributes of a GS change with its attached triangle's state as follows:
\begin{equation}
\begin{aligned}
\hat{\bm{\mu}}  &=  \lambda_1 \bm{x}_A  + \lambda_2 \bm{x}_B  + (1-\lambda_1-\lambda_2) \bm{x}_C +  z \vec{\bm{n}} ,  \\
\hat{\bm{r}} &= \bm{R} \bm{r}, \quad  \hat{\bm{s}}  = k \bm{s}
\end{aligned}
\label{eq_trans}
\end{equation}
where $x_A, x_B, x_C$ denote the world coordinates of the triangle's three vertices, 
$\vec{n}$ is the unit normal vector of this triangle,
and scaling factor $k$ is calculated as the ratio of a triangle's current area to its original area.
Note that the GS deforms with SMPL-X mesh's deformation in animation.

Different from~\cite{qian2024gaussianavatars}, we utilize two UV feature maps $(\mathcal{U}_c, \mathcal{U}_{\alpha})$ to store the RGB color
and opacity information of the 2DGS to facilitate later editing (Fig.~\ref{fig_app_logo}).
During rendering, the color and opacity information is sampled from the UV maps according to UV mapping and the barycentric coordinates ($\lambda_1, \lambda_2$).

Note that geometry attributes belonging to GS (i.e. $\lambda_1, \lambda_2, z$, orientation $\bm{r}$, and scale $\bm{s}$) are optimizable. 
In contrast, $\bm{x}_A, \bm{x}_B, \bm{x}_C$, orientations of the local coordinate's axes $\bm{R}$, normal vector $\vec{\bm{n}}$, and scaling scalar $k$ are calculated according to the transformed mesh vertices.

\subsection{Body generation}
\label{subsec:body}

We first generate a human body only in underwear represented by \modelname{}. 
As shown in Fig.~\ref{fig_overview}, the geometry branch and color branch are optimized alternatively by the RFDS loss calculated according to the given text prompt $T_1$.

Similar to previous work ~\cite{liao2024tada}, we utilize SMPL-X ~\cite{pavlakos2019expressive}, a deformable parametric 3D human body mesh model with predefined topology.
We add optimizable per-vertex displacement $\mathcal{D}$ to SMPL-X mesh $\mathcal{M}$ to obtain $\hat{\mathcal{M}}$.
To increase the density of 2DGS, we attach $(n^2/2)$ on-surface (i.e. 0 offsets along the normal vector) 2DGS uniformly distributed on every triangle.

\noindent\textbf{\textit{Geometry branch.}}
Optimizable parameters $\theta_1=\{\beta, \mathcal{D}\}$ are SMPL-X
shape $\beta$ and vertex displacement $\mathcal{D}$, which are transformed to a mesh $\hat{\mathcal{M}} = \mathcal{M}(\bm{\beta}) + \mathcal{D}$.
We fix other SMPL-X parameters ($\bm{\theta}$) and omit them for clarity.
Then $\hat{\mathcal{M}}$ is rendered as a normal image $\mathcal{N}$ for loss calculation.
\begin{equation}
\small
\nabla_{\theta_1}\mathcal{L}^{N}_{rfds}(\phi,\mathcal{N})=\mathbb{E}_{\epsilon,t}\bigg[w(t)(x-\epsilon-v_{\phi}(\mathcal{N}_{t},y,t))\frac{\partial \mathcal{N}}{\partial \theta_1 } \bigg]
\end{equation}

\noindent\textbf{\textit{Color branch.}} The optimizable parameters $\theta_2$ include the attributes $\theta_2 = \{\bm{u}, \bm{s}, \bm{r}, \mathcal{U}_c, \mathcal{U}_{\alpha}\}$ of 2DGS. 
Note that the geometry attributes ($\{\bm{u}, \bm{s}, \bm{r}\}$) of the 2DGS are also updated for local shape refinement.
According to Eq.~\ref{eq_gs} \& Eq.~\ref{eq_trans}  
, we render the RGB image $\mathcal{I}_b$ of $\mathcal{G}_{b}$ for loss calculation.
\begin{equation}
\small
\label{eq:RFIb}
\nabla_{\theta_2}\mathcal{L}^{I_b}_{rfds}(\phi,\mathcal{I}_b)=\mathbb{E}_{\epsilon,t}\bigg[w(t)(x-\epsilon-v_{\phi}({\mathcal{I}_b}_{t},y,t))\frac{\partial \mathcal{I}_b}{\partial \theta_2 } \bigg]
\end{equation}
Regularizations on the positions ($\mathcal{L}_{p}$), scales ($\mathcal{L}_{s}$), and rotations ($\mathcal{L}_{r}$) of the GS are applied to constrain the movement of the Gaussians, resulting in fewer artifacts~\cite{qian2024gaussianavatars}.
The total loss for the color branch is:
\begin{equation}
\label{eq:LGB}
\mathcal{L}_{\mathcal{G}_{b}} = \mathcal{L}^{I_b}_{rfds} + \lambda_p \mathcal{L}_{p} + \lambda_s\mathcal{L}_{s}+ \lambda_r\mathcal{L}_{r}
\end{equation}

\subsection{Cloth generation}
\label{subsec:cloth}

Thanks to the disentangled pipeline, we can sequentially generate garments, similar to the body generation process, with only one \textbf{\emph{crucial}} difference.
We introduce an additional mesh generation step since using a mesh with a predefined topology cannot handle clothes with drastically different topologies (e.g. dress vs trousers).

The garment, which is represented by the original 2DGS $\mathcal{G}_{m}$, is optimized using RFDS loss along with the human body $\mathcal{G}_{b}$ (fixed).
We propose a \emph{SAM-based filtering} operation to remove noisy Gaussians unrelated to the garment, facilitating human-garment separation.
When the optimization is finished, we extract its mesh using the TSDF algorithm~\cite{newcombe2011kinectfusion} from the rendered multiview depth image.
We fix the mesh topology of the generated garment after mesh simplification~\cite{garland1997surface} and bind 2DGS on it for later garment texture generation.

\vspace{5pt}
\noindent \emph{\textbf{2DGS based cloth generation for mesh extraction.}}
We first represent the garment using the original 2DGS representation (i.e. not bound to a mesh) for optimization so that we can generate garments of diverse types without topology constraints.
Specifically, we initialize a set of 2D Gaussians $\mathcal{G}_{m}$ from selected body regions to represent the garment (see \textbf{supplementary} for the initialization details). 
Unlike $\mathcal{G}_{b}$ in the body \modelname{}, the Gaussians $\mathcal{G}_{m}$ are not bound to the mesh, each assigned a color attribute $\bm{c}$ and opacity attribute $\bm{\alpha}$.
We merge $\mathcal{G}_{m}$  with the body $\mathcal{G}_{b}$  to render the clothed human image $\mathcal{I}_{a}$. 
Using RFDS loss $\mathcal{L}^{I_a}_{rfds}$ on $\mathcal{I}_{a}$, we only optimize $\mathcal{G}_{m}$ to align with the text prompt $T_2$.

To improve the geometry quality, we introduce two regularization terms.
First, we constrain a 2D Gaussian's distance to the mesh surface (i.e. point-to-surface distance) as:
\begin{equation}
\mathcal{L}_{dis} = ||\hat{\mu}-p_m||_2
\end{equation}
where $p_m$ is the projection point on the body mesh of a Gaussian's center $\hat{\mu}$.
Second, we apply a regularization on the rendered normal image $\mathcal{I}_n$ (of $\mathcal{G}_{m}$) and its Gaussian blurred image $G(\mathcal{I}_n)$, ensuring a smooth cloth surface:
\begin{equation}
\mathcal{L}_{smooth} = ||G(\mathcal{I}_n)-\mathcal{I}_n||^2_2
\end{equation}

\noindent \emph{\textbf{SAM-based filtering.}}
As shown in Fig.~\ref{fig_overview}, the generated $\mathcal{G}_{m}$ inevitably includes parts of the body.
To decouple the body and garment, we utilize SAM~\cite{kirillov2023segany} to filter out non-garment Gaussians.
Specifically, each Gaussian is assigned an \emph{extra} class attribute $o$ ($0$ for $\mathcal{G}_{b}$ and $1$ for $\mathcal{G}_{m}$ initially) to render a semantic image $\mathcal{I}_o$ with Eq.~\ref{eq_gs}.
We use SAM to obtain the semantic mask $\mathcal{M}$ (detailed in the \textbf{supplementary}) of the clothed human image $\mathcal{I}_{a}$ as the label and calculate the MSE loss $\mathcal{L}_{sam}$ between $\mathcal{M}$ and $\mathcal{I}_o$ to optimize $o$ of $\mathcal{G}_{m}$.
During $\mathcal{G}_{m}$ generation, we remove Gaussians whose $o$ are below 0.5 (i.e., non-garment 2DGS) every 500 iterations.

In summary, the optimizable parameters of $\mathcal{G}_{m}$ are $\{{\bm{u}, \bm{s}, \bm{r},\bm{c}, \bm{\alpha}, \bm{o}}\}$ and the full loss for optimization is:
\begin{equation}
\label{eq:LGF}
\mathcal{L}_{\mathcal{G}_{m}} = \mathcal{L}^{I_a}_{rfds} + \mathcal{L}_{sam} + \lambda_{dis} \mathcal{L}_{dis} +\lambda_{smooth} \mathcal{L}_{smooth}
\end{equation}

\noindent \emph{\textbf{Mesh extraction.}}
Following~\cite{huang20242d}, we reconstruct the garment mesh using the TSDF algorithm from multiview rendered depth images of $\mathcal{G}_{m}$.
We remove the garment's invisible faces inside the body mesh and simplify the mesh to $~10k$ faces through the mesh simplification algorithm~\cite{garland1997surface}, followed by Laplacian smoothing.
UV mapping can be obtained either automatically via the UV-Atlas~\cite{xatlas} or manually by defining cutting seams in modeling software~\cite{blender}.

\vspace{5pt}
\noindent \emph{\textbf{Texture generation.}}
We bind 2D Gaussians $\mathcal{G}_{g}$ to the extracted mesh and optimize them to generate the garment texture.
In this step, we add detailed material descriptions in the text prompt to generate the texture of the different materials (such as lace, denim, wool).
We obtain the garment-only image $\mathcal{I}_{g}$ by only rendering $\mathcal{G}_{b}$ and obtain the clothed human image $\mathcal{I}_{\dot{a}}$ by rendering $\mathcal{G}_{b}$ and $\mathcal{G}_{g}$ together.
We calculate RFDS losses $\mathcal{L}^{I_{g}}_{rfds}$ on $\mathcal{I}_{g}$  and $\mathcal{L}^{I_{\dot{a}} }_{rfds}$ on $\mathcal{I}_{\dot{a}}$ based on the respective text prompts, optimizing $\mathcal{G}_{m}$ while keeping $\mathcal{G}_{b}$ unchanged.
The optimizable parameters are $\bm{u}, \bm{s}, \bm{r}$, $\mathcal{U}_c$, $\mathcal{U}_{\alpha}$ of $\mathcal{G}_{g}$.
Similarly, we add regularization terms $\mathcal{L}_{p}, \mathcal{L}_{s}$, and $ \mathcal{L}_{r}$ to reduce the artifacts and the total loss is:
\begin{equation}
\label{eq:LGG}
\mathcal{L}_{\mathcal{G}_{g}} = \mathcal{L}^{I_{g} }_{rfds} + \mathcal{L}^{I_{\dot{a}} }_{rfds} + \lambda_p \mathcal{L}_{p} + \lambda_s\mathcal{L}_{s}+ \lambda_r\mathcal{L}_{r}
\end{equation}

\subsection{view-consistent refinement}
\label{subsec:refinement}

In this section, we propose view-consistent refinement, including a cross-view attention mechanism and a weighted denoising strategy based on a surface point's (viewing) incident angle to achieve consistent texture enhancement across views (Fig.~\ref{fig_overview}).
Directly using SDEdit to improve the texture as in~\cite{zhuang2024tip} often results in sharper but inconsistent results, as the lower-quality texture generated by SDS/RFDS loss requires adding stronger noise and more denoising steps, 
thereby increasing the cross-view inconsistency.

Our view-consistent refinement contains two critical improvements compared to a naive pixel-level refinement using SDEdit.
First, a cross-view attention mechanism ensures the consistency of texture style across views. 
Second, incident-angle-weighted denoising (IAW-DE) adjusts the pixel noise levels in the denoising process based on an incident-angle weight map, thereby focusing the texture refinement in regions that are ``better'' observed (i.e. closer to perpendicular to the camera) and thus more confident to update.
Specifically, we define a view sequence surrounding the object (body/garment) and progressively optimize its texture from each view. 
Starting from a predefined canonical view, we apply IAW-DE to enhance the texture image as the pseudo label to supervise the 2DGS rendered image.
This process is repeated for each view, with a cross-view attention mechanism to ensure a consistent texture style.

\vspace{0.1cm}
\noindent \emph{\textbf{Cross-view attention.}}
To maintain the consistent texture style across the views, we replace the self-attention in SD3 with cross-view attention during the denoising process inspired by video diffusion models~\cite{khachatryan2023text2video,yang2023rerender}.
We use the canonical and previous views ($\mathcal{V}_0, \mathcal{V}_{i-1}$) as the reference to maintain texture style consistency by concatenating their features into the calculation of key $\mathbf{K'} $ and value $\mathbf{V'}$.
\begin{equation}
\small
\mathbf{K'} = \mathbf{W}^K[z_{\mathcal{V}_0};z_{\mathcal{V}_{i-1}};z_{\mathcal{V}_i}]  \quad
\mathbf{V'} = \mathbf{W}^V[z_{\mathcal{V}_0};z_{\mathcal{V}_{i-1}};z_{\mathcal{V}_i}]
\end{equation}
where $\mathbf{W}^K$, $\mathbf{W}^V$ are the pre-trained parameters; $z_{(\cdot)}$ the latent feature from corresponding views.

\vspace{0.1cm}
\noindent \emph{\textbf{IAW-DE.}}
As~\cite{chen2023text2tex,cao2023texfusion,richardson2023texture} point out, the texture of a region should be determined by the view that observes it most directly.
Given a rendered view, we thus enhance textures in regions most directly observed, leaving less observed regions (e.g. boundaries) unchanged.
We use a weight map $\mathcal{W}_{i}$ (Fig.~\ref{fig_overview}, top-right) to measure the observation directness of each pixel in the $i$-th rendered image  $\mathcal{V}_{i}$.
Each pixel in $\mathcal{W}_{i}$ denotes the cosine similarity between the surface normal of point $p_j$ (i.e., the intersection of the pixel ray with the mesh) and the reversed view direction.
Since $p_j$ may be visible from multiple views, we aggregate its corresponding pixel values across all visible views and normalize them using temperature-scaled $Softmax$.
The weight map $\mathcal{W}_{i}$ is then resized to match the feature size in the diffusion model. 
According to $\mathcal{W}_{i}$, IAW-DE adds more noise to the higher-weighted feature pixels and performs more denoising iterations to obtain the refinement image $\hat{\mathcal{V}}_i$ (see Algorithm.1 in the \textbf{supplementary} for detail).
A weighted MSE loss is applied between the rendered image $\mathcal{V}_i$ and the refinement image $\hat{\mathcal{V}}_i$,
i.e., $\mathcal{L}_{ref} = ||\mathcal{W}_i (\hat{\mathcal{V}}_i-\mathcal{V}_i)||^2_2$.

\begin{figure*}
\centering
\includegraphics[width=1.0\textwidth]{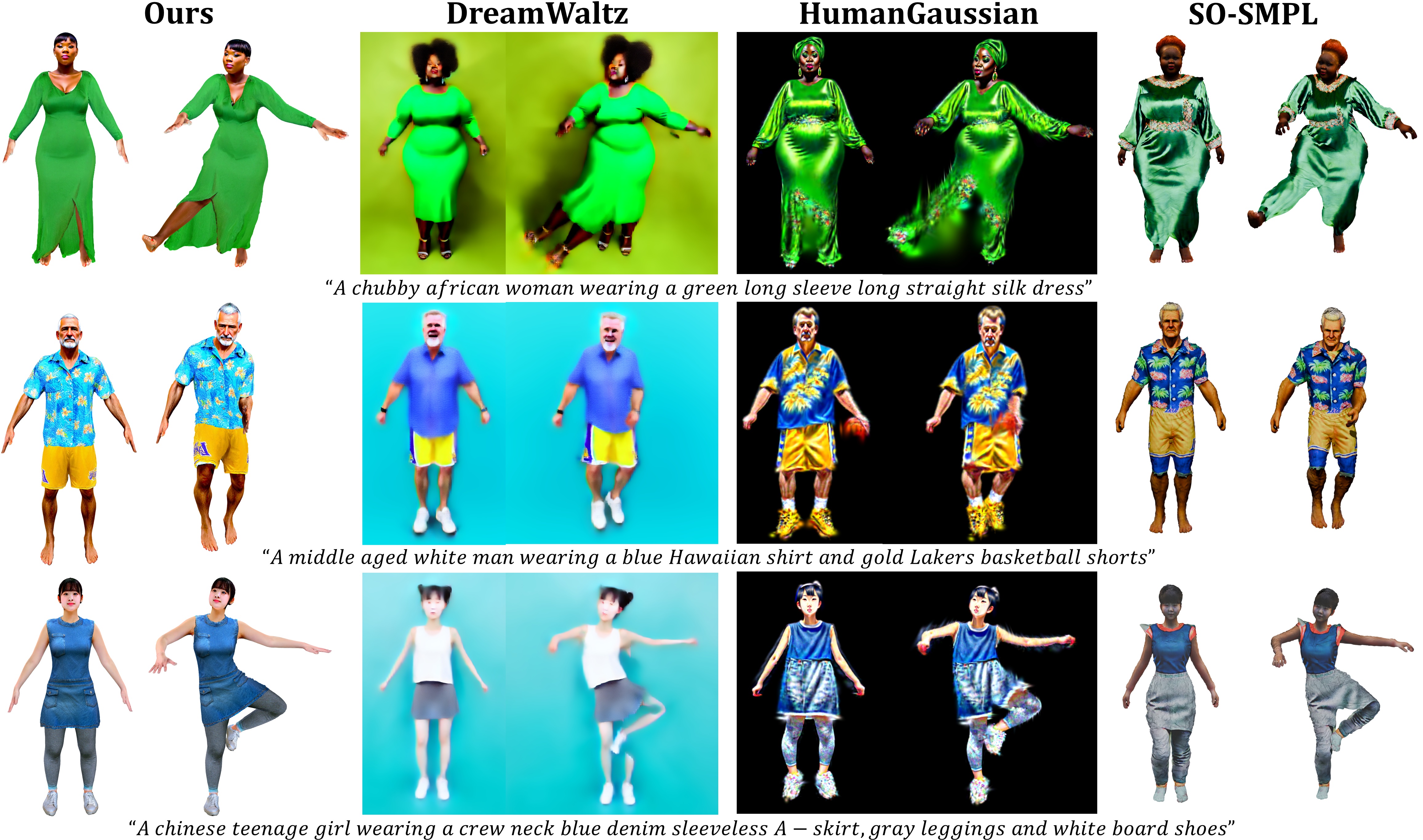}
\caption{
\textbf{Visual comparisons.}
Results generated by our method have significantly higher visual quality, accurately follow the input text prompt, and can be naturally animated.
In contrast, results from DreamWaltz are in low-resolution and contain obvious structural problems.
HumanGaussian produces unexpected results (e.g., basketball), with issues such as a split skirt in the animation. 
SO-SMPL is limited by its inability to generate clothing beyond the body's topology (e.g., dresses), restricting its applicability.
}
\label{fig:visual_compare}
\end{figure*}

\section{Experiments}

\noindent \emph{\textbf{Implementation Details.}}
We use the official code~\cite{huang20242d} to implement 2D Gaussian Splatting.
We set $\lambda_p$=$\lambda_s$=10, $\lambda_r$=1 in Eq.~\ref{eq:LGB}\&\ref{eq:LGG} and $\lambda_{dis}$=1, $\lambda_{smooth}$=100 in Eq.~\ref{eq:LGF}.
The rendered size is 1024×1024 and the RFDS loss is calculated on Stable Diffusion 3~\cite{esser2024scaling}.
All experiments are conducted on an NVIDIA A100 (40~GB).
The body generation takes 3K iterations, consuming roughly 60 minutes.
In the cloth generation, \emph{original 2DGS generation} and \emph{texture generation} require 2K iterations each, taking 30 and 20 minutes, respectively. 
View-consistent refinement takes 16 minutes, with 8 views around the object at 45\textdegree~intervals, and the 0\textdegree~view is the canonical view. 
More details (e.g., optimizer, learning rate) can be found in the \textbf{supplementary}.

\noindent \emph{\textbf{Baselines.}}
We compare our method with three baselines, including three representative 3D representations.
(1) Dreamwaltz~\cite{huang2024dreamwaltz}, which uses single NeRF~\cite{mildenhall2021nerf} representation and SDS loss for optimizing.
(2) HumanGaussian~\cite{liu2024humangaussian}, which models the whole clothed human with a set of 3DGS and using SD fine-tuned with depth maps as guidance.
and (3) SO-SMPL~\cite{wang2023disentangled}, a disentangled method that uses SDS-optimized traditional meshes (with predefined fixed topology) to represent body and garments.
All methods adopt SMPL-X~\cite{pavlakos2019expressive} as the body shape prior.
We implement all methods using their official codes.

\begin{table}[t]
\caption{Quantitative comparisons (rows 1-2) and user study (rows 3-5).
Human-G denotes HumanGaussian.
}
\small
\label{tab:Quantitative}
\renewcommand\tabcolsep{2pt}
\begin{tabular}{r|cccc}
\toprule
Metric  & DreamWaltz  & Human-G  & SO-SMPL & Ours \\
\midrule
\footnotesize{ViT-L/14} $\uparrow$ & 24.2 & 27.3 & 25.6 & \textbf{28.8}  \\
\footnotesize{ViTbigG-14} $\uparrow$ &  41.9  & 43.9 & 42.0  &  \textbf{45.8} \\
\midrule
\midrule
\footnotesize{Visual Quality} $\uparrow$ & 1.0\% & 8.4\% & 16.6\% & \textbf{74.0\%} \\
\footnotesize{Text Alignment} $\uparrow$ & 2.4\% & 9.2\% & 11.6\% & \textbf{76.8\%} \\
\footnotesize{Animation Realism} $\uparrow$ & 2.6\% & 8.6\% & 20.4\% & \textbf{68.4\%} \\
\bottomrule
\end{tabular}
\end{table}

\subsection{Comparisons with Baselines}

\myparagraph{Qualitative comparisons.}
Fig.~\ref{fig:visual_compare} shows visual comparisons of avatars generated by our method and the baselines at the same pose.
Obviously, our method generates higher texture quality results, more accurate adherence to the input text prompt, and more realistic animation (e.g., the deformation of the green dress).
In contrast, due to the limitations of the 3D representation and training pipelines, baselines generate suboptimal results and encounter challenges in animation.

DreamWaltz and HumanGaussian results usually do not follow the provided text prompts accurately, possibly because they generate all the clothes at once from much more complicated text descriptions.
In contrast, we only need to provide detailed descriptions for generating the current cloth.
For example, HumanGaussian generates unintended basketballs and sneakers (Fig.~\ref{fig:visual_compare}, row 2), while DreamWaltz inaccurately generates white vests (row 3).
Moreover, DreamWaltz uses implicit NeRF, resulting in low-resolution results (due to slow rendering) and broken body integrity in animation.
Results from HumanGaussian fail to maintain cloth integrity during the animation (e.g., tearing of dresses) since the garment 3DGS are also attached to the SMPL body mesh for movement.
SO-SMPL can not generate clothes (e.g., dresses) beyond SMPL-X's topological structure, limiting its applicability.
See the animation videos of all methods in our \textbf{supplementary} for clearer comparisons.

\noindent \emph{\textbf{Quantitative comparisons.}}
We adopt CLIP~\cite{radford2021learning} similarity to assess the alignment of 10 generated avatars with their corresponding text prompts.
We render 120 images per avatar in different views and poses. 
Specifically, the elevation angle ranges from 0° to 30°, while the azimuth angle ranges from -60° to 60.
The poses include A-pose and random dance pose.
Tab.~\ref{tab:Quantitative} shows the CLIP similarity scores using two backbones, i.e. ViT-L/14~\cite{radford2021learning} and ViTbigG-14~\cite{ilharco_gabriel_2021_5143773}. 
These results demonstrate the superiority of our method in both CLIP metrics, indicating that our generated avatars align better with the text prompts.

\noindent \emph{\textbf{User study.}}
We conduct a user study with 50 participants to evaluate the 10 generated avatars according to three aspects, including visual quality, alignment to the text prompts, and animation quality.
We adopt the same dance pose sequence to drive each method and show the rendered animation videos to the participants (see our \textbf{supplementary}).
As shown in Tab.~\ref{tab:Quantitative}, our method outperforms the baselines with a significant margin, receiving 74\% votes for visual 
 quality, 76.8\% for alignment, and 68.4\% for animation quality.

\subsection{Ablation Studies}
\label{sec_ablation}

\begin{figure}
\includegraphics[width=1.0\columnwidth]{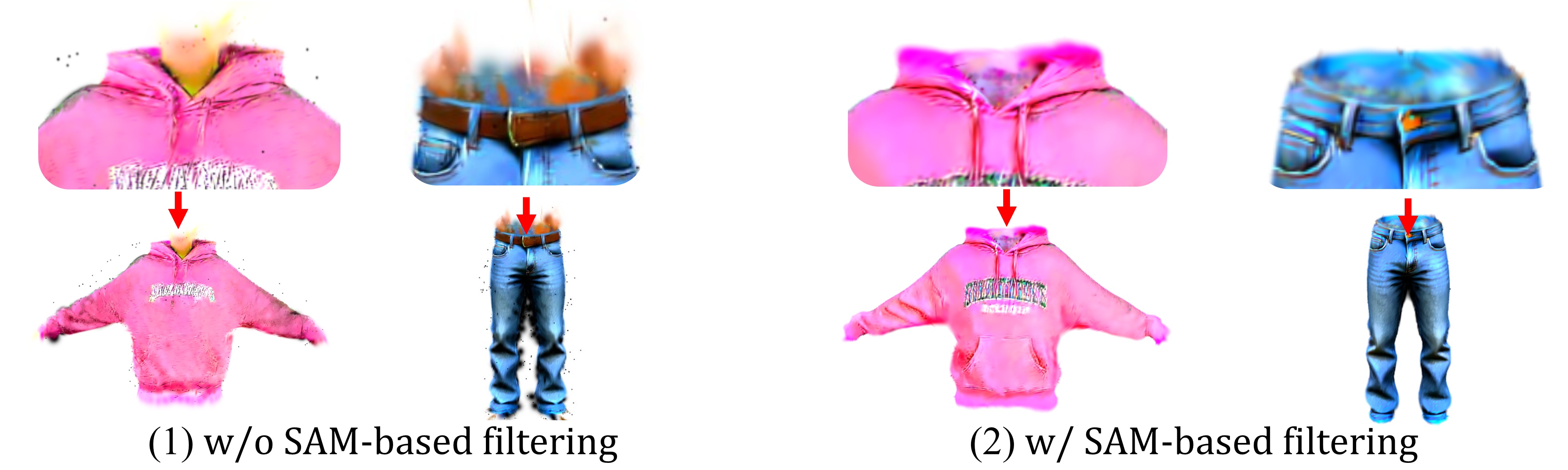}
\caption{
\textbf{Ablation study on SAM-based filtering.} The results demonstrate its effectiveness in filtering out the noise belonging to the body, achieving better body-garment separation.
}
\label{fig_abl_sam}
\end{figure}

\noindent \emph{\textbf{Ablation studies on SAM-based filtering.}}
Fig.~\ref{fig_abl_sam} demonstrates the benefit of SAM-based filtering in cloth generation (Sec.\ref{subsec:cloth}).
Without SAM-based filtering, the generated garment 2DGS always contains obvious noise from its adjacent part (e.g., body parts indicated by the arrow).
Introducing SAM-based filtering effectively handles the noise, resulting in better body-garment separation.

\begin{figure*}
\centering
\includegraphics[width=1.0\textwidth]{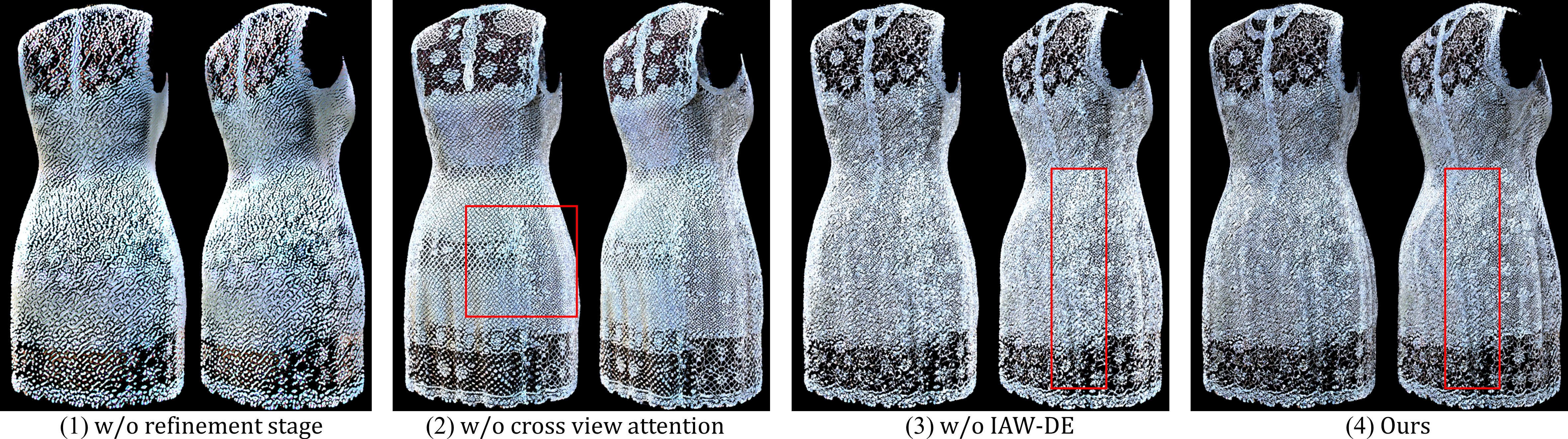}
\caption{
\textbf{Ablation study on view-consistent refinement.}
To better demonstrate the texture, we render a single layer of the white lace dress (i.e., the layer not occluded by the body) on a black background.
Our view-consistent refinement (4) effectively improves the suboptimal textures generated by RFDS loss (1).
This refinement includes two key components: cross-view attention, which ensures texture style consistency across views (2 vs. 4);
and IAW-DE, which reduces the blurriness of the texture caused by view inconsistency (3 vs. 4).
}
\label{fig_abl_ref}
\end{figure*}

\noindent \emph{\textbf{Ablation studies on the view-consistent refinement.}}
We conduct ablative experiments to verify the effects of view-consistent refinement and its key components: cross-view attention and IAW-DE.
To better demonstrate the texture details, we render a single-layer white lace dress (i.e., the layer not occluded by the body) on a black background.
Fig.~\ref{fig_abl_ref} (1)\&(4) show the suboptimal textures generated by the RFDS loss and the improved textures by our view-consistent refinement, respectively.
Our view-consistent refinement significantly enhances the texture quality while maintaining the consistency of the texture style.

Fig.~\ref{fig_abl_ref} (2)\&(3) demonstrate the benefit of cross-view attention and IAW-DE in view-consistent refinement.
Without cross-view attention, the consistency of the texture decreases after enhancement.
Without IAW-DE, the overlapping texture regions of two rendered views can easily get blurred even if their textures are slightly inconsistent.
IAW-DE allows each view to focus on enhancing the most directly observed regions, thereby reducing texture blurriness caused by view inconsistency.

\begin{figure}
\centering
\includegraphics[width=1.\columnwidth]{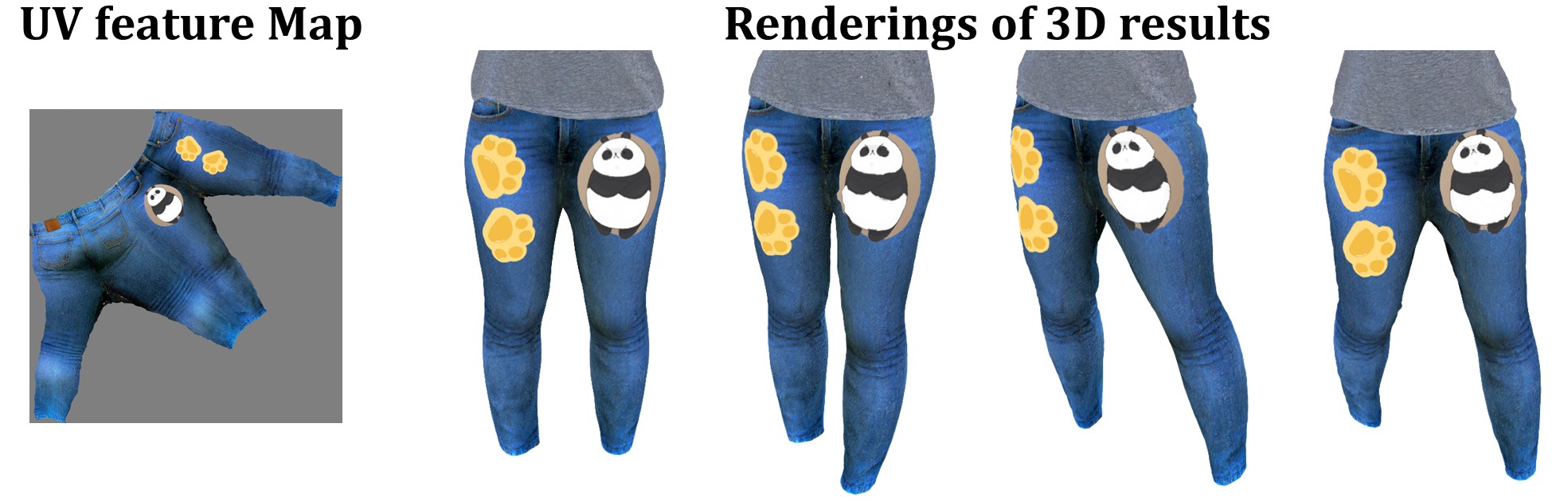}
\caption{
\textbf{Manual texture editing} by modifying $\mathcal{U}_c$.
}
\label{fig_app_logo}
\end{figure}

\subsection{Applications}

\noindent \emph{\textbf{Clothing replacement and Animation.}}
As presented in Fig.~\ref{fig:teaser}, our method allows the users to alter the clothes of the generated avatars without changing the body appearance, which is an essential feature for virtual try-on applications.
In animation, we use SMPL-X parameters from AIST++~\cite{li2021learn} to drive the body mesh and simulate garment mesh deformation in Blender~\cite{blender}.
Fig.~\ref{fig:teaser}\&~\ref{fig:visual_compare} show our avatars in different poses, demonstrating realistic garment-body interactions due to our decoupled mesh-based representation.
Videos are provided in the \textbf{supplementary}.

\noindent \emph{\textbf{Manual texture editing.}}
Thanks to the UV color maps $\mathcal{U}_c$, we can directly modify $\mathcal{U}_c$ to edit the model's texture.
In Fig.~\ref{fig_app_logo}, we add a logo to the cloth by copying it into $\mathcal{U}_c$. 
The pattern in the rendered image is clear and deforms naturally in cloth animation.

\begin{figure}
\centering
\includegraphics[width=1.\columnwidth]{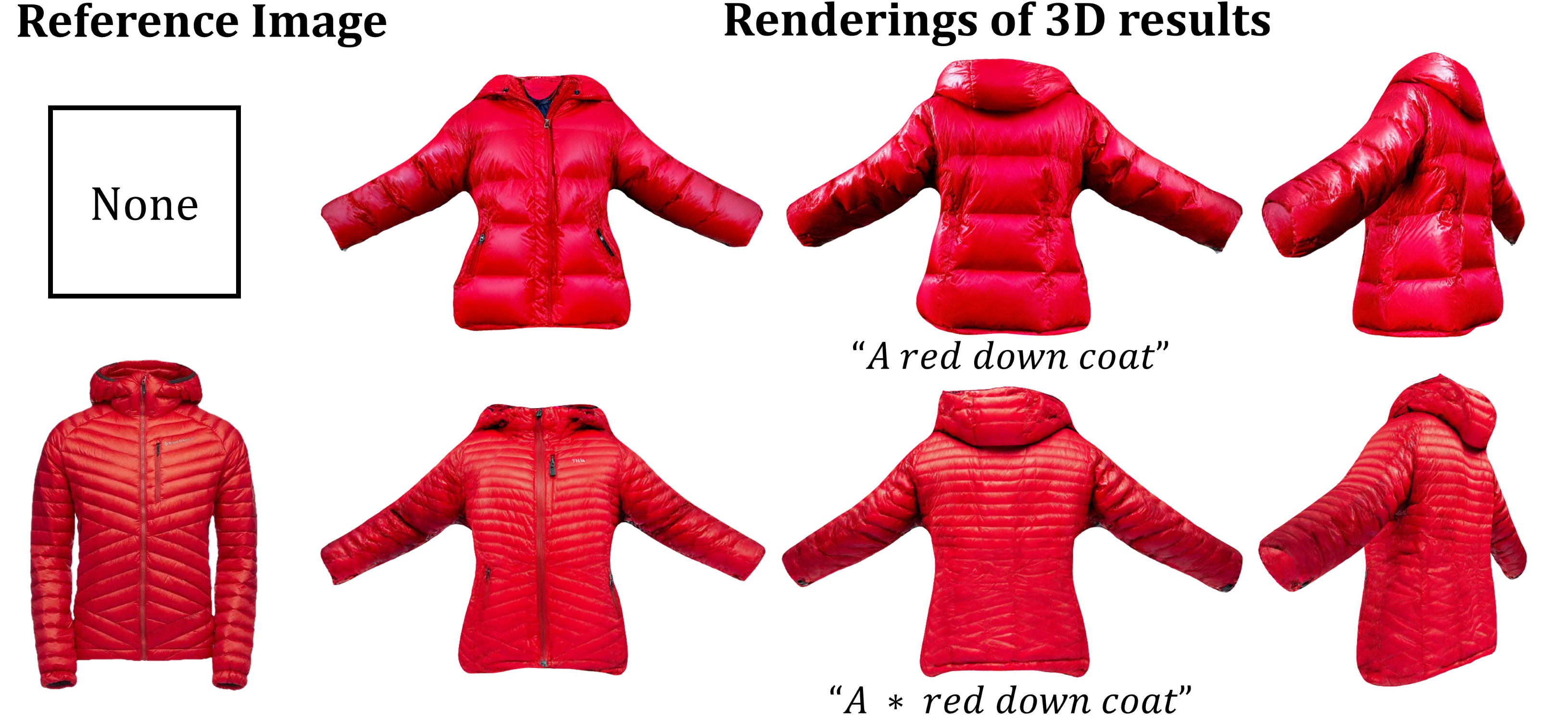}
\caption{ 
\textbf{Reference image} guided generation for better appearance control.
``$*$" denotes the special token.
}
\label{fig_app_ref}
\end{figure}

\noindent \emph{\textbf{Generating with a reference image.}}
Our approach also supports more accurate and user-friendly appearance control via a reference image as a prompt.
Using TIP-Editor~\cite{zhuang2024tip}, we encode the reference image as a special token and integrate it into the text prompt.
As shown in Fig.~\ref{fig_app_ref}, compared to only using text prompts, the generated clothes with the reference image exhibit high consistency with the reference, significantly improving appearance control.

\section{Conclusion and Limitations}

In this paper, our proposed \sysname{} allows the users to generate digital humans with decoupled bodies and garments from text prompts. 
We separately generate the human body and garments represented by different \modelname{}s, which is a hybrid representation binding 2DGS with a mesh.
Thanks to our disentangled generation pipeline and \modelname{} representation, \sysname{} naturally enables clothing replacement and transparent fabric generation, and supports clothing simulation in animation.
Extensive experiments demonstrate that \sysname{} outperforms existing methods, generating higher visual quality, supporting more features, and enabling more realistic animations.
One limitation of \sysname{} is that the TSDF algorithm for mesh extraction only reconstructs the garment surface, omitting internal structures like pockets. 

\section*{Acknowledgement}
This work is supported in part by the National Key R\&D Program of China under Grant No.2024YFB3908503, in part by the National Natural Science Foundation of China under Grant NO.~62322608 and NO.~62325605.

{
    \small
    \bibliographystyle{ieeenat_fullname}
    \bibliography{main}
}

\clearpage
\setcounter{page}{1}
\setcounter{figure}{7}
\setcounter{equation}{11}
\setcounter{section}{0}
\setcounter{table}{1}
\renewcommand\thesection{\Alph{section}}
\maketitlesupplementary
\begin{figure*}
\centering
\includegraphics[width=0.65\textwidth]{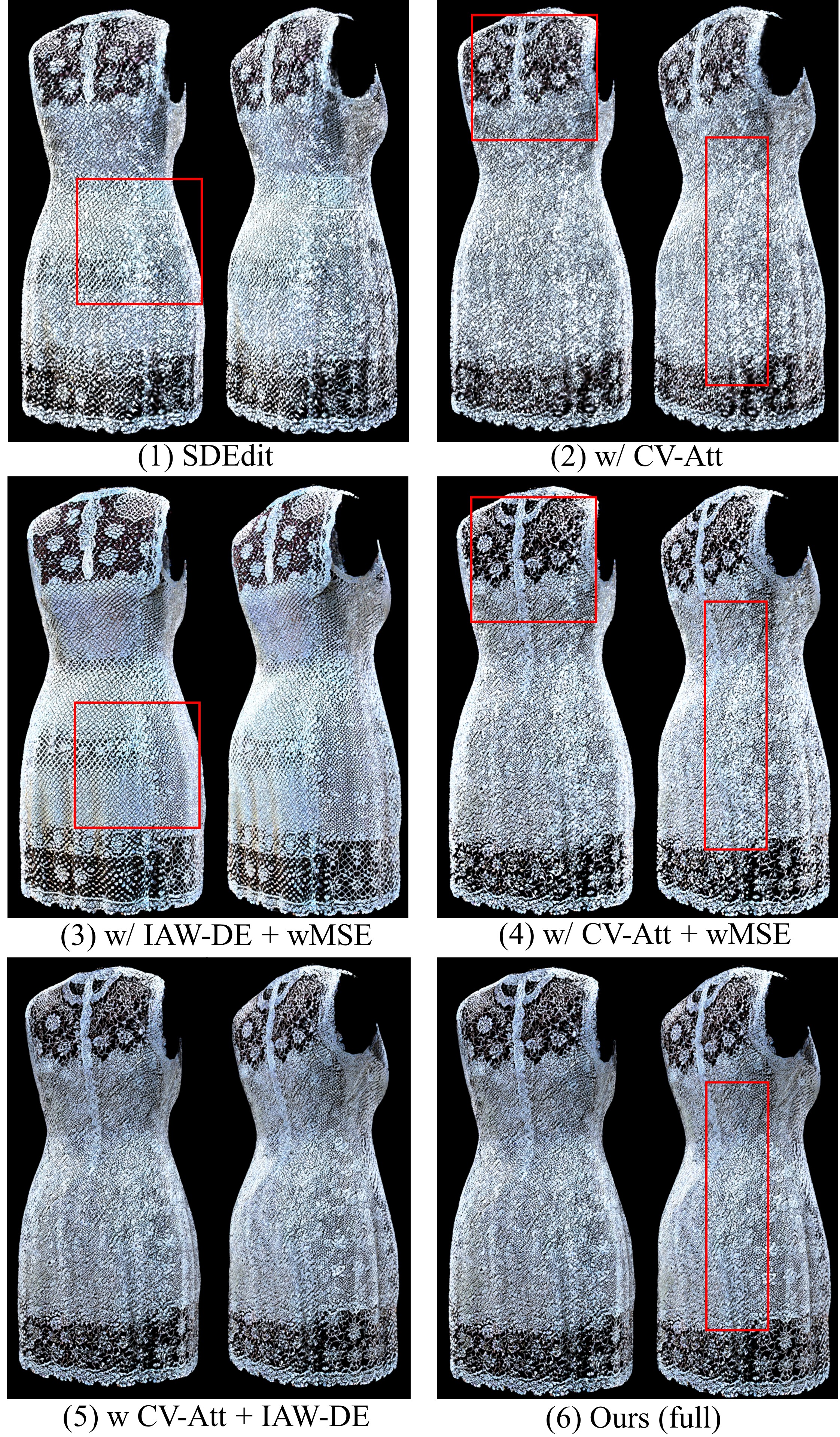}
\caption{
Ablation study on three components in view-consistent refinement: cross-view attention (CV-Att), IAW-DE, and using a weighted MSE loss with weighted map $\mathcal{W}$ (wMSE).
Using the original SDEdit (1) results in inconsistent texture styles and thus blurred overlapping regions due to view inconsistency. 
Removing cross-view attention (3) reduces texture consistency, proving the importance of cross-view attention in maintaining texture style consistency.
Without IAW-DE and the weighted MSE loss (2), the overlapping texture regions of two rendered views can easily get blurred even if their textures are only slightly inconsistent.
Applying a weighted MSE loss (4) partially alleviates the blurred texture in overlapping regions caused by view inconsistency.
However, our IAW-DE addresses this issue more effectively (4 vs. 5) by allowing each view to focus on enhancing the most directly observed regions.
Although the weighted MSE loss offers limited improvement (5 vs. Ours), we also include it for its implementation simplicity.
}
\label{sup_abl_ref}
\end{figure*}

In this document, we provide more results of ablation studies (\ref{sec_ablation}) and implementation details of our method (\ref{sec_details}).

\section{More ablation studies} \label{sec_ablation}

\subsection{Ablation study on view-consistent refinement.} 
We conduct an ablation study to demonstrate the benefits of three components in the view-consistent refinement step: cross-view attention (CV-Att), IAW-DE, and using a weighted MSE loss with weighted map $\mathcal{W}$ (wMSE).
To better demonstrate the texture, we render a single layer of the white lace dress (i.e., the layer not occluded by the body) on a black background.

As shown in Fig.~\ref{sup_abl_ref}(1), directly using original SDEidt as in~\cite{zhuang2024tip}, the results have inconsistent texture style.
Moreover, the overlapping texture regions of two rendered views are blurred due to view inconsistency.
Without cross-view attention (Fig.~\ref{sup_abl_ref}(3)), the consistency of the texture decreases after enhancement, proving the importance of cross-view attention in maintaining texture style consistency.
Without IAW-DE and the weighted MSE loss (Fig.~\ref{sup_abl_ref}(2)), the overlapping texture regions of two rendered views can easily get blurred even if their textures are slightly inconsistent.
Fig.~\ref{sup_abl_ref} (4) shows that applying a weighted MSE loss partially alleviates the blurred texture in overlapping regions caused by view inconsistency.
However, IAW-DE addresses this issue more effectively (4 vs. 5) by allowing each view to focus on enhancing the most directly observed regions.
Although the weighted MSE loss offers limited improvement (Fig.~\ref{sup_abl_ref}(5)), we also include it for its implementation simplicity.

\subsection{Ablation study on different 3D representations.} 

\begin{figure*}
\centering
\includegraphics[width=0.8\textwidth]{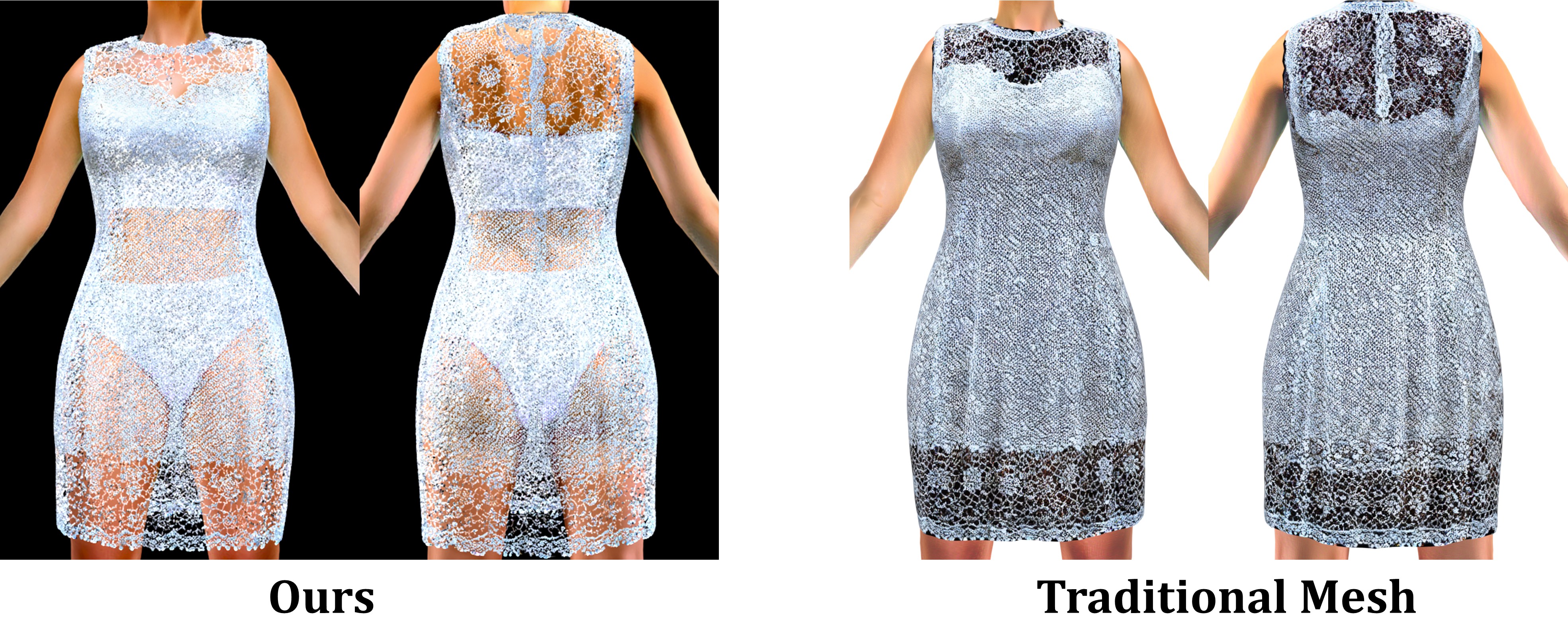}
\caption{
Ablation study on different 3D representations to show the advantage of GSM in handling transparent fabric texture.
We compare our \modelname{} with the traditional mesh representation to generate a white lace dress while keeping all the other settings the same.  
Our \modelname{} can handle transparent fabric texture while traditional mesh fails to learn the transparent texture using existing differentiable engine (e.g. Nvdiffrast \cite{Laine2020diffrast}) automatically.
}
\label{sup_3d}
\end{figure*}

In Fig.~\ref{sup_3d}, we compare our \modelname{} with the traditional mesh representation to generate a white lace dress while keeping all the other settings the same.  
As demonstrated in Fig.~\ref{sup_3d}, our \modelname{} can handle transparent fabric texture while traditional mesh with texture cannot automatically learn texture transparency.

\section{Implementation Details} \label{sec_details}

We provide more implementation details that cannot be included in the main paper due to space.

\begin{algorithm}[t]	
\setlength{\itemsep}{-1.0cm}	
\caption{ Incident-angle-weighted denoising (IAW-DE)}	
\label{al_IAW_DE}
{
{\bf Input:} The weight map $\mathcal{W}$; the rendered image $\mathcal{V}$; the text prompt $y$; refine strength $n$ (i.e., number of denoising steps)\\
{\bf Output:} The refined image \: $\hat{\mathcal{V}}$. 

\begin{algorithmic}[1]	
    \STATE $z=$ SD3\_encode($\mathcal{V}$)
    \STATE $\hat{z}_n=$ addnosie($z, n$), $z_n=$ denosie($\hat{z}_n, n, y$)
    \STATE \textbf{for} $ t = n-1$ to $0$ \textbf{do}
    \STATE \quad  $\hat{z}_t=$ addnosie($z, t$)
    \STATE \quad \textbf{for} $\mathcal{W}_j$ in $\mathcal{W}$ \textbf{do}
    \STATE \quad \quad \textbf{if} $\mathcal{W}_j > \frac{t}{n} $ \textbf{then}
    \STATE \quad \quad \quad $\hat{z}_{t,j}=z_{t+1,j}$
    \STATE \quad \quad \textbf{end if}
    \STATE \quad \textbf{end for}
    \STATE \quad $z_t=$ denosie($\hat{z}_t, t, y$)
    \STATE \textbf{end for}
    \STATE $\hat{\mathcal{V}} = $ SD3\_decode($z_0$)
    \STATE \textbf{return} $\hat{\mathcal{V}}$
\end{algorithmic}
}
\end{algorithm}

\subsection{Details of IAW-DE}

As described in Algorithm.~\ref{al_IAW_DE}, 
we input the weight map $\mathcal{W}$, the rendered image $\mathcal{V}$, the rendered image $\mathcal{V}$, the text prompt $y$, and refine strength $n=0.4$ to Stable Diffusion 3 (SD3)~\cite{esser2024scaling}, obtaining the refined image $\hat{\mathcal{V}}$ via incident-angle-weighted denoising (IAW-DE).

\begin{table*}[t]
\centering
\small
\caption{Selected body regions to initialize the garment 2DGS.}
\label{tab_sleceted_region}
\renewcommand\tabcolsep{2pt}
\begin{tabular}{c|c}
\toprule
Types  & Selected regions \\
\midrule
t-shirt & Spines, Shoulders, Arms  \\
long-shirt & Spines, Shoulders, Arms, ForeArms  \\
hoodie & Spines, Shoulders, Arms, ForeArms  \\
down coat & Spines, Shoulders, Arms, ForeArms, Hips  \\
coat & Spines, Shoulders, Arms, ForeArms, Hips, UpLegs  \\
shots &  Hips, UpLegs  \\
pants &  Hips, UpLegs, Legs  \\
shoes &  Foots, ToeBases  \\
sleeveless dress &  Spines, Shoulders, Hips, UpLegs  \\
long sleeve long dress &  Spines, Shoulders, Arms, ForeArm, Hips, UpLegs, Legs  \\
\bottomrule
\end{tabular}
\end{table*}

\begin{figure}
\centering
\includegraphics[width=0.9\columnwidth]{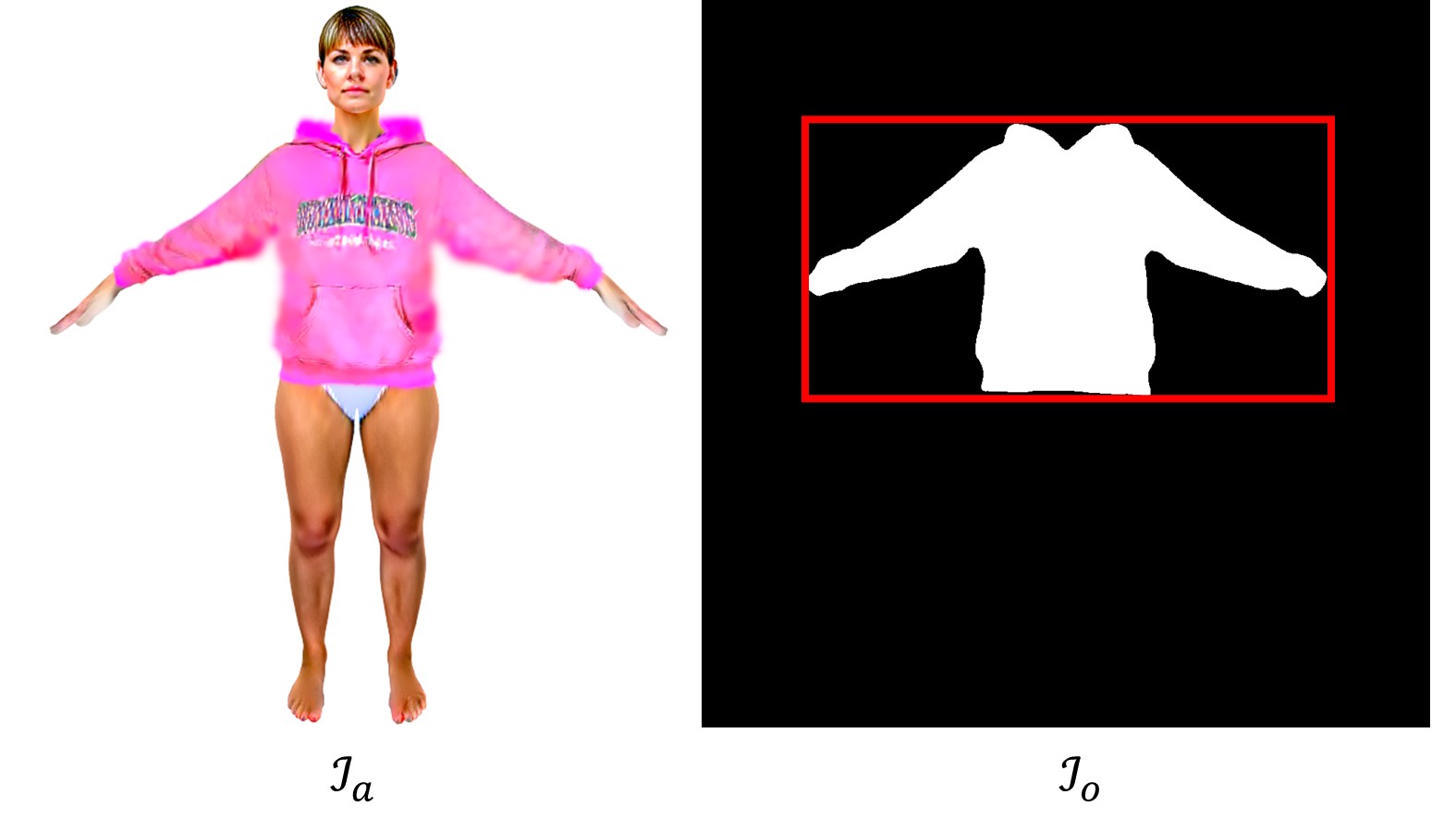}
\caption{Inputs to SAM: the clothed human image $\mathcal{I}_{a}$ and the bounding box (red) of the Garment 2DGS (i.e. the hoodie) as SAM's prompt.
}
\label{sup_sam}
\end{figure}

\subsection{Formula of the regularizations in Eq.~6}

In body generation (Sec.~\ref{subsec:body}) and texture generation (Sec.~\ref{subsec:cloth}), we add regularizations on the positions ($\mathcal{L}_{p}$), scales ($\mathcal{L}_{s}$), and rotations ($\mathcal{L}_{r}$) into Eq.~6 and Eq.~10 to constrain the movement of the Gaussians.

Positions regularization $\mathcal{L}_{p}$ constrains the projection point of the Gaussian center on the bound triangle to be within the boundaries of the triangle, with a height offset below 0.1:
\begin{equation}
\begin{aligned}
\mathcal{L}_{p} = \mathcal{L}_{\lambda}(\lambda_1) +& \mathcal{L}_{\lambda}(\lambda_2) + \mathcal{L}_{\lambda}(1-\lambda_1-\lambda_2)  + ||\text{max}(z, 0.1))||^2_2 
\\
& \mathcal{L}_{\lambda} = \begin{cases}
-\lambda, \quad \text{if}\;\lambda < 0 
\\
0, \quad \text{if} \;0< \lambda < 1  
 \\
\lambda-1, \quad \text{if}\; \lambda > 1
\end{cases}
\end{aligned}
\end{equation}

Following GaussianAvatars~\cite{qian2024gaussianavatars}, we constrain the local scale of each 2D Gaussian to remain below 0.6:
\begin{equation}
\mathcal{L}_{p} = || \text{max}(\bm{s}, 0.6)||^2_2
\end{equation}

Rotations regularization $\mathcal{L}_{r}$ constrains the normal direction of the 2D Gaussian disk to be consistent with the normal direction $\vec{\bm{n}}$ of the bound triangle:
\begin{equation}
\mathcal{L}_{r} =  \text{cosine}(\bm{R} \bm{r} \bm{n}, \vec{\bm{n}})
\end{equation}
where $\bm{n}$ is vector $[0, 0, 1]$.

\subsection{Details of initializing 2D Gaussians in Sec.~\ref{subsec:cloth}}

As shown in Tab.~\ref{tab_sleceted_region}, we use SMPL-X~\cite{pavlakos2019expressive} part segments to select the 2DGS from the corresponding body regions to initialize the garment 2DGS for different clothing types.

\subsection{Details of using SAM to obtain semantic masks}

We implement SAM~\cite{kirillov2023segany} via Huggingface~\cite{wolf2019huggingface}. Using the bounding box (the red box in Fig.~\ref{sup_sam}) of the rendered garment's 2DGS as the box prompts, we input it along with the clothed human image $\mathcal{I}_{a}$ to SAM to obtain the semantic mask $\mathcal{M}$ of the garment.

\begin{table}[t]
\centering
\caption{Learning rates of different parameters.}
\label{tab_lr}
\renewcommand\tabcolsep{10pt}
\begin{tabular}{l|c}
\toprule
Parameter  & Learning rate \\
\midrule
$\beta$ & 0.01  \\
$\mathcal{D}$ & 0.0001  \\
$\bm{u}$ & 0.0002$\to$0.00002  \\
$\bm{s},\bm{r}$ & 0.005  \\
$\mathcal{U}_c, \bm{c}$ & 0.01  \\
$\mathcal{U}_{\alpha}, \bm{\alpha}, \bm{o}$ & 0.1  \\
\bottomrule
\end{tabular}
\end{table}

\subsection{More implementation details}

We use the official code~\cite{huang20242d} to implement 2D Gaussian Splatting.
We use the Adam optimizer with $beta1=0.9$, $beta2=0.999$, $weight decay=0$, and $epsilon=10^{-15}$ for optimization and the learning rates of different parameters can be found in Tab.~\ref{tab_lr}.
We set $\lambda_p$=$\lambda_s$=10, $\lambda_r$=1 in Eq.6\&10 and $\lambda_{dis}$=1, $\lambda_{smooth}$=100 in Eq.9.
The rendered size is 1024×1024 and the RFDS loss is calculated on Stable Diffusion 3~\cite{esser2024scaling}, with the CFG weight 100 and time steps $t\sim \mathcal{U} (0.02, 0.98)$.
All experiments are conducted on an NVIDIA A100 (40~GB).
The body generation takes 3K iterations, consuming roughly 60 minutes.
In the cloth generation, \emph{original 2DGS generation} and \emph{texture generation} require 2K iterations each, taking 30 and 20 minutes, respectively. 
In body generation and cloth generation, the view angles range from -15\textdegree to 30\textdegree in elevation and -180\textdegree to 180\textdegree in azimuth.
View-consistent refinement takes 16 minutes, with 8 views around the object at 45\textdegree~intervals, and the 0\textdegree~view is the canonical view.

\end{document}